\documentclass{article}
\pdfoutput=1

\usepackage[letterpaper,left=1.5in,right=1.5in,top=1in,bottom=1in]{geometry}
\usepackage{times}
\usepackage{natbib}
\usepackage{amsmath,amssymb}
\usepackage{amsmath,amsfonts,bm}

\def\eqref#1{equation~\ref{#1}}
\def\1{\bm{1}}

\DeclareMathAlphabet{\mathsfit}{\encodingdefault}{\sfdefault}{m}{sl}
\SetMathAlphabet{\mathsfit}{bold}{\encodingdefault}{\sfdefault}{bx}{n}

\usepackage{hyperref}
\usepackage{url}
\usepackage{enumitem}
\usepackage{graphicx}
\usepackage{booktabs}
\usepackage{longtable}

\setcitestyle{authoryear,round,citesep={;},aysep={,},yysep={;}}

\makeatletter

\renewcommand{\normalsize}{\fontsize{10}{11}\selectfont}
\renewcommand{\small}{\fontsize{9}{10}\selectfont}
\renewcommand{\footnotesize}{\fontsize{9}{10}\selectfont}

\renewcommand{\large}{\fontsize{12}{14}\selectfont}

\renewcommand{\LARGE}{\fontsize{17}{20}\selectfont}

\normalsize

\def\section{\@startsection {section}{1}{\z@}{-2.0ex plus -0.5ex minus -.2ex}%
  {1.5ex plus 0.3ex minus0.2ex}{\large\sc\raggedright}}
\def\subsection{\@startsection{subsection}{2}{\z@}{-1.8ex plus -0.5ex minus -.2ex}%
  {0.8ex plus .2ex}{\normalsize\sc\raggedright}}
\def\subsubsection{\@startsection{subsubsection}{3}{\z@}{-1.5ex plus -0.5ex minus -.2ex}%
  {0.5ex plus .2ex}{\normalsize\sc\raggedright}}
\def\paragraph{\@startsection{paragraph}{4}{\z@}{1.5ex plus 0.5ex minus .2ex}%
  {-1em}{\normalsize\bf}}
\def\subparagraph{\@startsection{subparagraph}{5}{\z@}{1.5ex plus 0.5ex minus .2ex}%
  {-1em}{\normalsize\sc}}

\skip\footins 9pt plus 4pt minus 2pt
\def\footnoterule{\kern-3pt \hrule width 12pc \kern 2.6pt }

\leftmargini\leftmargin \leftmarginii 2em
\belowdisplayskip \abovedisplayskip
\def\@maketitle{\vbox{\hsize\textwidth \centering
  {\LARGE\sc \@title\par}
  \vskip 0.25in
  {\@author}
  \vskip 0.3in minus 0.1in}}

\makeatother

\renewenvironment{abstract}{\vskip.075in\centerline{\large\sc
Abstract}\vspace{0.5ex}\begin{quote}}{\par\end{quote}\vskip 1ex}

\title{A Statistical Audit of Physical-AI Benchmark Redundancy}

\author{%
  \begin{tabular}[t]{c}\bf Zaruhi Navasardyan\\ Metric AI Lab\\ \texttt{zaruhi@metricailab.com}\end{tabular}
  \hspace{0.5in}
  \begin{tabular}[t]{c}\bf Hrant Davtyan\\ Metric AI Lab\\ \texttt{hrant@metricailab.com}\end{tabular}
}
\date{}

\begin{document}

\maketitle
\thispagestyle{plain}
\pagestyle{plain}

\begin{abstract}
Physical AI models are evaluated on suites of benchmarks that
differ across model reports, leaving the model-by-benchmark matrix sparse and
the relationship between benchmarks unmeasured. We construct a matrix of 51
models on 12 physical AI benchmarks, selected from a registry of 51 benchmarks
and 152 models by reporting density, combining scores from model cards and
benchmark papers with our own evaluation runs under each benchmark's official
protocol. We measure how much information the benchmarks share and show quantitative evidence of Redundancy. Redundancy affects reported rankings:
collapsing the two substitute pairs into single columns moves 22 of 51 models
by three or more places under an equally weighted average. We then select
benchmarks greedily under a utility combining score dispersion with variance
not explained by the already-selected set, and obtain a four-benchmark subset
retaining 78.5\% of the utility of all 12, on which we fit a Bradley--Terry
ranking. The procedure requires only benchmark-level scores with sufficient
overlap and is not specific to physical AI.

\vspace{0.3cm}
\noindent
\centering
\small
\textbf{Project home:} \href{https://metric-ai-lab.github.io/metabench/}{https://metric-ai-lab.github.io/metabench/} \quad

\end{abstract}

\section{Introduction}

Physical AI is young enough that its evaluation has no common ground. Vendors
report each new model on a hand-picked set of benchmarks, and the sets barely
overlap: the model $\times$ benchmark matrix implied by public reports is
mostly empty. Two consequences follow. First, models cannot be compared, 
there is no shared axis, no MTEB-style leaderboard, on which competing systems
line up. Second, the numbers that \emph{are} reported can mislead: a headline
score is typically an average over the chosen benchmarks, and if those
benchmarks measure overlapping abilities, the average silently double-counts
the shared signal.

The overlap is not hypothetical. Consider pointing - outputting a location to
identify an object, find free space, or resolve a referring expression. This
single capability is routinely evaluated several times within one report:
Gemini Robotics-ER 1.5 reports Point-Bench, RefSpatial, RoboSpatial-Pointing,
and Where2Place \citep{gemini_robotics_2025};
RoboBrain 2.0 evaluates RoboSpatial, RefSpatial-Bench, and Where2Place
\citep{robobrain_2025}; Qwen3-VL reports RefSpatial, RoboSpatial-Home \citep{qwen3_vl_2025}. The same repetition appears in 3D layout
reasoning and relational question answering. A recent survey catalogs over 45
spatial-reasoning benchmarks and finds coverage heavily concentrated in
relational-static questions and sparse elsewhere \citep{liu2025spatial}.

New benchmarks appear for good reasons: older ones saturate, leak into
pre-training data, or admit shortcuts. But each new benchmark is introduced
on the premise that it measures something the existing ones do not, and this
premise is not checked. To date, no one has quantified how much unique signal
a physical-AI benchmark adds beyond the benchmarks already in use.

We treat this as a measurement problem. We assemble a dense matrix of 51 models
on 12 physical-AI benchmarks combining scores from official model cards and
benchmark papers with our own evaluation runs under each benchmark's official
protocol. On this matrix we ask two questions, each targeting one
of the frictions above:

\begin{description}
  \item[RQ1 - Redundancy.] How much information do the 12 benchmarks share,
        and how strongly does this redundancy inflate pooled averages?
  \item[RQ2 - Sufficiency.] How small can a benchmark suite be while still
        separating models and covering the non-redundant abilities of the
        full suite?
\end{description}

Statistical auditing of this kind exists for text LLMs. Metabench
\citep{metabench_2024} fits item response theory to 28{,}632 items from six LLM
benchmarks across ${>}5{,}000$ models and shows that under 3\% of items suffice
to reconstruct full scores; \citet{burnell2023revealing} factor-analyze 29 LLMs
over 27 HELM tasks and recover three capability factors explaining 82\% of
variance. These audits, however, require dense item-level
response data --- a setting unavailable in physical AI, where results are
published as benchmark-level aggregates. We show that a meaningful audit is
possible at the benchmark level, and that it answers both questions. The twelve
benchmarks carry far less independent information than their count suggests:
the closest substitutes agree at $\rho = 0.88$, and the median benchmark has
roughly half its variance reconstructible from the other eleven. That redundancy
is not inert --- collapsing just the two substitute pairs moves $22$ of $51$
models by three or more places, so part of a model's standing reflects how often
the suite happens to measure what it is good at. It is also compressible: four
benchmarks retain $78.5\%$ of the suite's discriminating power.

Our contributions are as follows:

\begin{enumerate}
  \item \textbf{A dense evaluation matrix for physical AI.} Scores for 51
        models on 12 physical AI benchmarks, assembled from model cards and
        benchmark papers and completed by our own runs under each benchmark's
        official protocol, plus 9 general benchmarks for the same models. This
        makes previously non-comparable models directly comparable and enables
        the audit below.
  \item \textbf{Redundancy (RQ1).} We quantify shared and unique information
        across the 12 benchmarks, identify close substitutes, and show how
        averaging correlated benchmarks inflates pooled scores --- a standard
        practice in vendor reporting.
  \item \textbf{Sufficiency (RQ2).} We give selection criteria --- sharp model
        separation, distance from saturation, information beyond the selected
        set --- and forward-select a 4-benchmark suite that preserves the
        non-redundant signal of the full suite. We use it to produce a Bradley--Terry ranking.
\end{enumerate}

\section{The Benchmark--Model Matrix}
\label{sec:matrix}

We start our analysis by indexing 51 physical AI benchmarks assembled from model
cards, papers, and official blogs. Considering our
analysis objectives, we apply three criteria for selection.
\textbf{(i) Density.} A benchmark has to be reported for at least 5 candidate
models to ensure overlapping scores for covariance analysis.
\textbf{(ii) Recency.} Each benchmark recurs across recent model reports, so the
suite reflects what the field actually uses to make claims.
\textbf{(iii) Diversity.} Together they span the key tasks measured by
physical-AI benchmarks, so a finding of redundancy cannot be attributed to
having picked twelve versions of the same task.

Table~\ref{tab:benchmarks} provides the final list of the 12 benchmarks that
satisfy these criteria (out of 51), including the ability each claims to
measure, its task format, its size, and how many of the models have a score on
it. The benchmark definitions and evaluation splits follow the introducing
sources: \textsc{VSI-Bench}~\citep{yang2024vsi},
\textsc{EmbSpatial-Bench}~\citep{du2024embspatial},
\textsc{RefSpatial-Bench}~\citep{zhou2025roborefer},
\textsc{Where2Place}~\citep{yuan2024robopoint},
\textsc{ERQA}~\citep{gemini_robotics_2025_original},
\textsc{CV-Bench}~\citep{tong2024cambrian},
\textsc{SAT}~\citep{ray2024sat},
\textsc{RoboSpatial}~\citep{song2024robospatial},
\textsc{RealWorldQA}~\citep{xai2024grok},
\textsc{OmniSpatial}~\citep{jia2025omnispatial},
\textsc{MindCube}~\citep{wang2025mindcube}, and
\textsc{BLINK}~\citep{fu2024blink}.
Ten of them are multiple-choice; the remaining two
(\textsc{RefSpatial-Bench} and \textsc{Where2Place}) instead require the model
to emit image coordinates, and a prediction is scored correct when the point
falls inside a target mask. All 12 report model performance on a 0--100 scale.

\begin{table}[t]
\centering
\caption{The 12 physical-AI benchmarks. \emph{Items} = number of samples in the benchmark; \emph{$n$} = models with a score in our matrix, of 51. \emph{$g$} = Gini coefficient.}
\label{tab:benchmarks}
\vspace{0.5em}
\setlength{\tabcolsep}{4pt}
\small
\begin{tabular}{llrrrrrr}
\toprule
\textbf{Benchmark} & \textbf{Claimed ability} &
\textbf{Items} & \textbf{$n$} & 
\textbf{Mean} & \textbf{SD} & \textbf{Min--Max} & \textbf{$g$} \\
\midrule
\textsc{VSI-Bench}        & Visual-spatial intelligence (video) & 5{,}130  & 51 & 46.9 & 12.8 & 12.6--69.5 & 0.153 \\
\textsc{EmbSpatial}       & Egocentric spatial relations        & 3{,}640  & 51 & 73.2 & 8.0  & 43.2--84.1 & 0.057 \\
\textsc{RefSpatial-Bench} & Spatial referring (pointing)        & 200      & 51 & 29.9 & 18.1 & 0.3--72.2  & 0.343 \\
\textsc{Where2Place}      & Affordance pointing / free space    & 100      & 50 & 42.3 & 19.8 & 7.6--76.0  & 0.265 \\
\textsc{ERQA}             & Embodied reasoning, planning        & 400      & 49 & 44.5 & 8.5  & 25.7--65.0 & 0.104 \\
\textsc{CV-Bench}         & Classical CV as VQA (depth, count)  & 2{,}638  & 49 & 81.7 & 6.1  & 61.0--89.2 & 0.039 \\
\textsc{SAT}              & Dynamic spatial aptitude            & 150      & 49 & 68.6 & 11.7 & 45.3--88.0 & 0.097 \\
\textsc{RoboSpatial}      & Robot-centric spatial reasoning     & 350      & 47 & 50.5 & 9.5  & 29.4--72.6 & 0.103 \\
\textsc{RealWorldQA}      & Real-world spatial QA               & 765      & 42 & 68.0 & 9.0  & 40.6--80.4 & 0.069 \\
\textsc{OmniSpatial}      & Comprehensive spatial cognition     & 1{,}533  & 42 & 46.4 & 6.1  & 26.5--59.6 & 0.071 \\
\textsc{MindCube}         & Spatial mental models (multi-view)  & 21{,}154  & 42 & 42.2 & 11.7 & 18.7--69.2 & 0.153 \\
\textsc{BLINK}            & Multi-image visual perception       & 3{,}807  & 41 & 65.4 & 12.3 & 43.8--86.3 & 0.106 \\
\bottomrule
\end{tabular}
\end{table}

As can be seen from Table~\ref{tab:benchmarks}, the benchmark means span from 30
to 82 points. If several benchmarks measure the same skill yet differ in
difficulty, their raw scores will differ in level and spread but not in ranking
models identically. Thus, a correlation computed on ranks will be unaffected,
yet any analysis in score units would confound difficulty with information.
Therefore, we z-score every benchmark column before multivariate steps, and
separately keep Spearman rank correlations as the default pairwise measure.
Difficulty itself is not discarded, it re-enters as a selection criterion in
Section~\ref{sec:minimal}, where a benchmark near its ceiling is penalized
regardless of what it measures.

In terms of diversity, we design the matrix by selecting benchmarks that cover diverse tasks and settings. We classify them into 5 groups. \emph{Pointing} benchmarks (\textsc{RefSpatial-Bench},
\textsc{Where2Place}) ask for a location that satisfies a referring expression
or an affordance --- the format closest to what a robot policy consumes.
\emph{Single-image relational} benchmarks (\textsc{EmbSpatial},
\textsc{CV-Bench}, \textsc{OmniSpatial}, \textsc{RealWorldQA})
show one view and ask about relative position, depth, count, or the scene from
another viewpoint. \emph{Multi-view and video} benchmarks (\textsc{MindCube}, \textsc{SAT},
\textsc{VSI-Bench}) cannot be answered from a single frame: MindCube supplies
two to four views of one scene, VSI-Bench a walkthrough video, and both require
integrating evidence across them. \emph{Embodied} benchmarks (\textsc{ERQA},
\textsc{RoboSpatial}) frame questions from a robot's point of view: what can be
grasped, where an object may be placed, whether a configuration is feasible.
\textsc{BLINK} stands apart as a \emph{general visual perception} benchmark.
We use this grouping only as the taxonomy a reader
would expect, not as base for grouping them during the analysis.

Similarly, we index 152 models with at least one physical-AI benchmark score. We
then use the final list of 12 benchmarks to select the 51 models from the
registry that have at least two thirds (8 of 12) of the scores reported. The
final list includes models released between 2024 and 2026, from 15 different
providers, both open-weight and closed, ranging from 1B to 241B in size
(counted for open-weight models only). While some of the models are generalist
VLMs, others are specifically trained for robotics or spatial tasks. Those
models are usually post-trained on a named open base model, allowing comparison
before and after domain post-training. The full list of models is described in 
Appendix~\ref{app:models}.

We extract scores from model cards and papers. However, published reporting
alone leaves the matrix too sparse for a covariance analysis, so we run the
missing evaluations. For all models we use each benchmark's official evaluation
code and prompts whenever available, with greedy decoding and the benchmark's
own answer-parsing rule. This contributes 159 additional data points to the
matrix. We did not conduct a systematic reproduction study: our runs targeted
missing scores, and while we validated our implementation on models with
published results, we do not claim to have verified the published part of the
matrix.

\section{Redundancy (RQ1)}
\label{sec:redundancy}

A benchmark suite is informative only when its benchmarks provide distinct
evidence about model capabilities. Redundancy arises when adding a benchmark
contributes little new information. Beyond unnecessary evaluation cost, such
redundancy can also give disproportionate weight to capabilities measured
repeatedly when benchmark scores are aggregated.

\begin{figure}[t]
    \centering
    \includegraphics[width=0.49\linewidth]{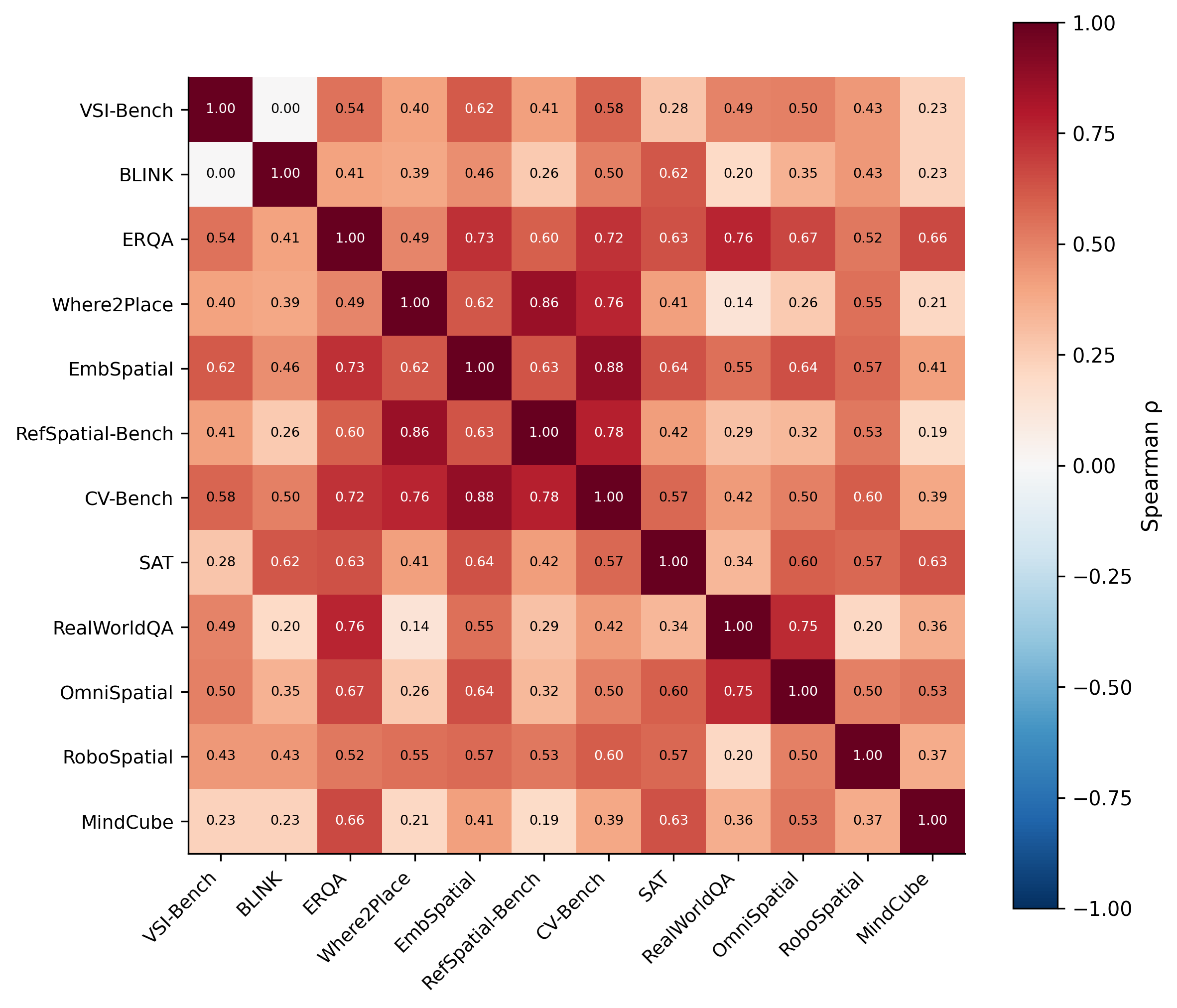}
    \hfill
    \includegraphics[width=0.49\linewidth]{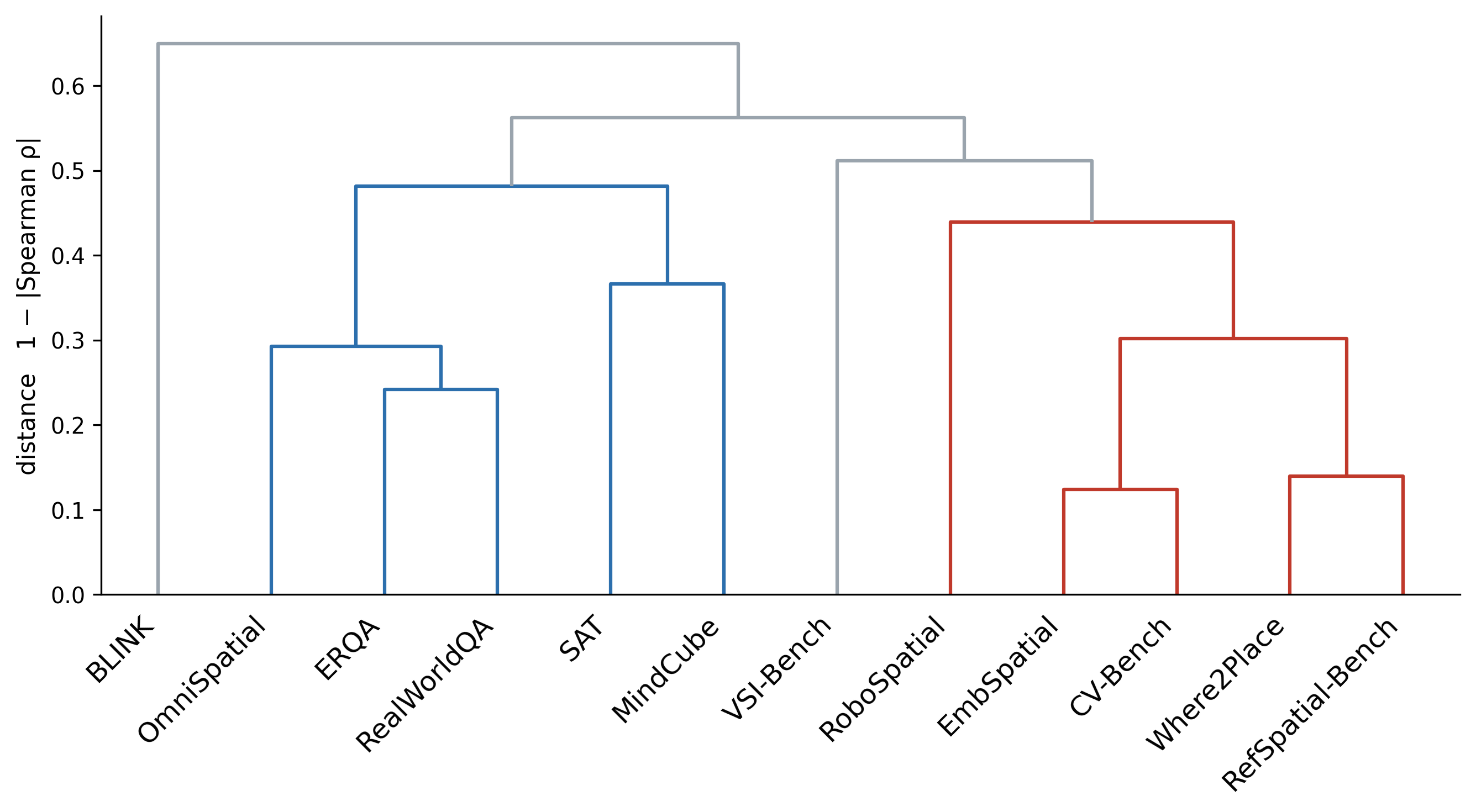}
    \caption{Left: pairwise Spearman $\rho$ across the 12 benchmarks, computed
    on pairwise-complete rows. Right:
    average-linkage hierarchical clustering under $D = 1 - \rho$.}
    \label{fig:rq1-heatmap}
\end{figure}

We start our redundancy analysis by computing Spearman pairwise correlations. The rank correlations
allow us to measure how similarly two benchmarks rank models, independent of
benchmark difficulty, which is inherently present in absolute score values. The
average pairwise $\rho$ is $0.487$, with all correlations being positive.
The full correlation matrix is available in Figure~\ref{fig:rq1-heatmap}.
Additionally, the figure shows the dendrogram from hierarchical clustering on
the benchmark--model matrix using $1 - \rho$ as the distance metric. The
dendrogram itself does not add new information; rather, it serves as an easy
way to visually observe the benchmark groupings.

Analyzing Figure~\ref{fig:rq1-heatmap} shows that two pairs stand out as
substitutes with Spearman correlation above $0.8$:
\textsc{EmbSpatial}\,$\leftrightarrow$\,\textsc{CV-Bench} ($\rho = 0.876$, 95\%
CI $[0.78, 0.93]$, $n = 49$) and
\textsc{Where2Place}\,$\leftrightarrow$\,\textsc{RefSpatial-Bench} ($\rho =
0.860$, $[0.73, 0.93]$, $n = 50$).

\textsc{BLINK} appears to be the most unique benchmark in terms of rank
correlations, forming an individual branch in the dendrogram. Other notable
pairs are \textsc{ERQA}\,$+$\,\textsc{RealWorldQA} at $\rho = 0.758$ and
\textsc{RealWorldQA}\,$+$\,\textsc{OmniSpatial} at $0.748$.

We dive deeper into the redundancy analysis and study how reconstructable a
benchmark is from all of its peers, rather than from a single one. To do that,
we apply ridge regression to predict each benchmark from the
other 11 and compute the leave-one-out cross-validated $R^2$: at each
iteration, we hold a single model out and calculate its prediction from a model
trained on the remaining model scores. The lower the LOO $R^2$, the more unique
the benchmark.

Table~\ref{tab:loo} reports the results of the regression. The benchmarks exhibit substantial variation in redundancy. Where2Place, RefSpatial-Bench, and ERQA are the most predictable from the remaining suite ($R^2$=0.727, 0.721, and 0.706), indicating considerable overlap in the evidence they provide. Their predictor sets are also strongly interconnected, with these three benchmarks repeatedly predicting one another. This agrees with our findings from pairwise analysis, where \textsc{Where2Place} and \textsc{RefSpatial-Bench} formed the highest Spearman-correlated pair. MindCube emerges most often (7 times out of 11) among the strongest predictors of other benchmarks, with ERQA and RefSpatial-Bench next (5 times each), suggesting that they act as hubs of shared benchmark behavior. In contrast, RealWorldQA, RoboSpatial, and BLINK are substantially less predictable ($R^2$=0.319, 0.351, and 0.378). We attribute this to the distinct information that those benchmarks add to the rest of the benchmarks, yet we acknowledge that a low $R^2$ can also be caused by noise in the benchmark and the resulting measurement error. The remaining benchmarks occupy an intermediate regime. Thus, benchmark redundancy is not one-dimensional: highly redundant benchmarks may nevertheless be valuable as representatives of shared capability structure, whereas highly unique benchmarks provide complementary evidence. This distinction is important when constructing a compact suite, where benchmark selection should balance unique information against coverage of shared structure.

\begin{table}[t]
\centering
\caption{Leave-one-model-out predictability of each benchmark from its eleven
physical peers. $1 - \text{LOO } R^2$ is the
benchmark's uniqueness. Top 3 predictors are the benchmarks which contribute most to explaining current benchmark's variance.}
\label{tab:loo}
\vspace{0.5em}
\small
\begin{tabular}{lrrrl}
\toprule
\textbf{Benchmark} & \textbf{$n$} & \textbf{LOO $R^2$} & \textbf{Uniqueness $1-R^2$} & \textbf{Top 3 predictors} \\
\midrule
\textsc{Where2Place}      & 50 & 0.727 & 0.273 & RefSpatial-Bench, MindCube, BLINK \\
\textsc{RefSpatial-Bench} & 51 & 0.721 & 0.279 & Where2Place, ERQA, MindCube \\
\textsc{ERQA}             & 49 & 0.706 & 0.294 & MindCube, RefSpatial-Bench, RealWorldQA \\
\textsc{CV-Bench}         & 49 & 0.614 & 0.386 & EmbSpatial, ERQA, RefSpatial-Bench \\
\textsc{MindCube}         & 42 & 0.599 & 0.401 & ERQA, RefSpatial-Bench, SAT \\
\textsc{SAT}              & 49 & 0.468 & 0.532 & OmniSpatial, MindCube, BLINK \\
\textsc{EmbSpatial}       & 51 & 0.463 & 0.537 & CV-Bench, OmniSpatial, Where2Place \\
\textsc{OmniSpatial}      & 42 & 0.453 & 0.547 & EmbSpatial, SAT, RealWorldQA \\
\textsc{VSI-Bench}        & 51 & 0.421 & 0.579 & ERQA, MindCube, BLINK \\
\textsc{BLINK}            & 41 & 0.378 & 0.622 & RefSpatial-Bench, MindCube, VSI-Bench \\
\textsc{RoboSpatial}      & 47 & 0.351 & 0.649 & VSI-Bench, MindCube, RealWorldQA \\
\textsc{RealWorldQA}      & 42 & 0.319 & 0.681 & ERQA, OmniSpatial, RoboSpatial \\
\bottomrule
\end{tabular}
\end{table}

To substantiate our claim that aggregated measurement is misleading under a correlated benchmark suite, we compute the arithmetic average score per model and rank models from highest to lowest. Weighting 12 benchmarks equally weights an ability in proportion to how many times the suite happens to measure it (in our case, pointing receives $2/12$ of the weight while video-spatial reasoning gets $1/12$).

We then use the insights from the redundancy analysis to collapse each substitute pair into a single benchmark. Specifically, we replace each of the two pairs that RQ1 flags as substitutes (\textsc{RefSpatial-Bench}/\textsc{Where2Place}, \textsc{EmbSpatial}/\textsc{CV-Bench}) with the mean of its two columns, leaving ten columns: eight untouched benchmarks and two collapsed abilities. The arithmetic average and the ranking are recomputed accordingly. We observe that among 51 models, 22 change their positions by 3 or more places. For example, \textsc{MiMo-Embodied-7B} drops 9 places and \textsc{Gemini Robotics-ER 1.5} drops 8. Similarly, models that are strong elsewhere and relatively weak at pointing rise: \textsc{GPT-4o} gains 9 places and \textsc{Claude-Sonnet-4} gains 8. We do not propose the de-duplicated ranking as the correct leaderboard --- collapsing pairs is itself a choice. However, this comparison isolates how much of a model's position is an artifact of the suite's composition rather than of its capability.

This section shows that redundancy exists, 
Appendix~\ref{app:structure} goes one step further and examines what the shared variance consists of: a single
principal component explains 55.2\% of the suite's variance and tracks general
vision--language capability, estimated from nine non-physical benchmarks on the same
models, at $\rho = 0.95$; residualizing every benchmark on that external axis roughly
halves the mean pairwise correlation, from $0.487$ to $0.250$. Roughly half of what the
12 benchmarks share is therefore general capability rather than anything specific to
physical AI.

\section{A Minimal Benchmark Suite (RQ2)}
\label{sec:minimal}

Section~\ref{sec:redundancy}
establishes that the 12 benchmarks share most of their signal, which implies
that some subset of them reproduces most of the evidence the full suite
provides. It does not tell us which subset, or how small it can be. However, informativeness is not a property a benchmark holds on
its own: it depends on which benchmarks are already included in the suite. A benchmark that
would be indispensable in isolation is worthless next to a substitute, as the
pointing pair of Section~\ref{sec:redundancy} demonstrates. Therefore, this section builds the minimal benchmark suite greedily: benchmarks are added one at a time based on their marginal gain conditional on already selected benchmarks in the suite.

We require two properties of a benchmark before it earns a place in the suite.
First, it must separate models. A benchmark that assigns nearly the same score
to every model orders them by noise, and contributes nothing to a leaderboard
however distinct the ability it measures. Second, it must carry information the
selected benchmarks do not. A benchmark that is reconstructable from the
selected set adds no evidence.
We score each candidate $b$ against the selected
set $S$ as the product of the two:
\begin{equation}
\label{eq:utility}
U(b \mid S) = a(b)\,\bigl(1 - R^2(b \sim S)\bigr).
\end{equation}

We take the product rather than a weighted sum because the two properties are
not substitutes: a benchmark that fails either one is not worth running, and
the product sends its utility to zero, whereas a sum would let a high value on
one term compensate for a near-zero value on the other.

\textbf{Discrimination $g(b)$} measures how widely a benchmark spreads models. We use the
Gini coefficient of the benchmark score distribution across models
(Table~\ref{tab:benchmarks}). A benchmark with high $g$ separates models
sharply; one with low $g$ scores them all alike.

\textbf{Marginal information $1 - R^2(b \sim S)$} is the share of $b$'s variance the
selected set cannot already reproduce. $R^2(b \sim S)$ is the fit of an OLS regression of
$b$ on all of $S$.

\subsection{Selected Benchmark Suite}
\label{sec:rq3_suite}

\begin{figure}[t]
    \centering
    \includegraphics[width=0.62\linewidth]{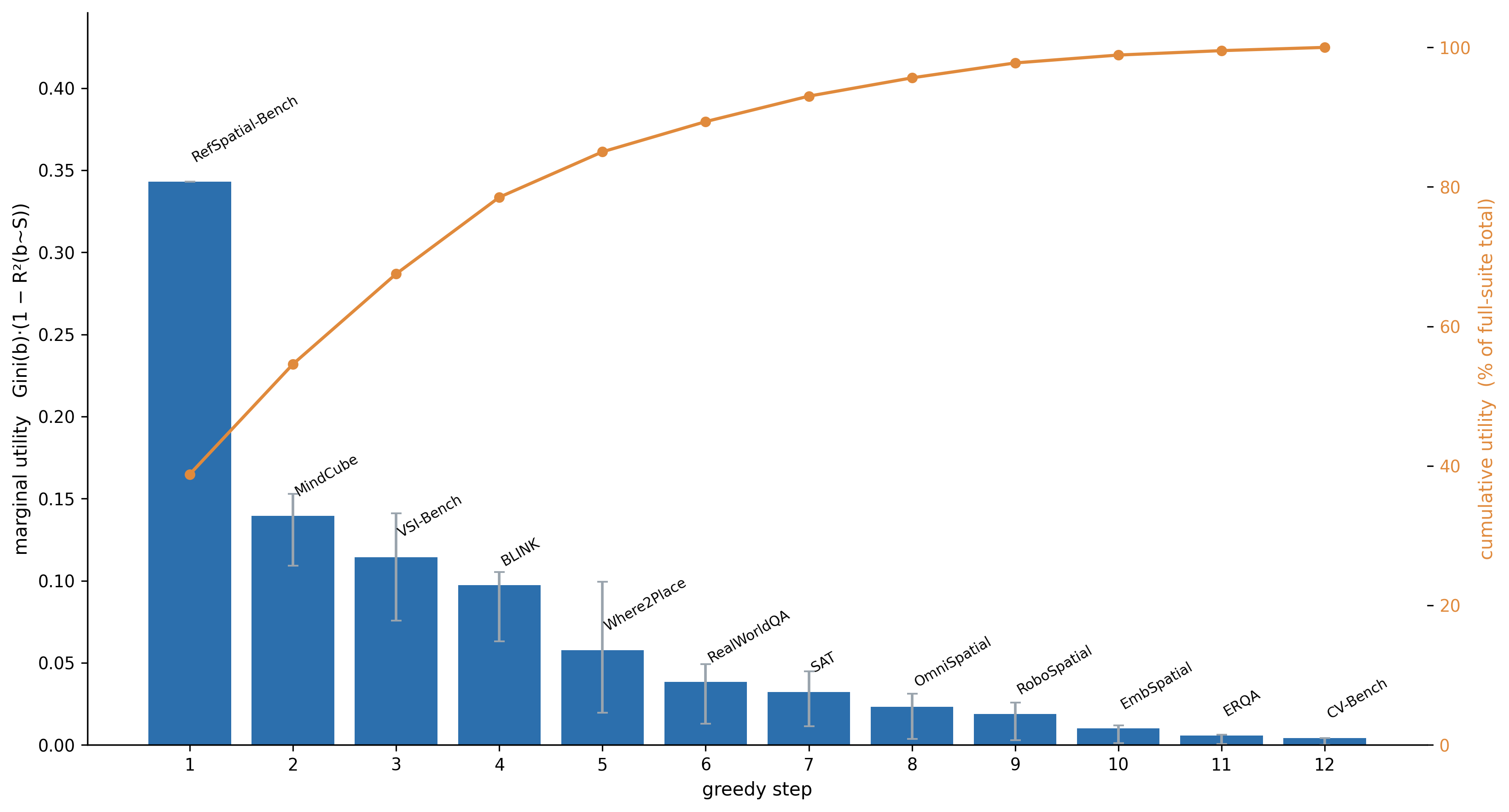}
    \caption{Forward-selection path. Bars are the per-step marginal utility
    $U(b \mid S) = g(b)\,(1 - R^2(b \sim S))$; the line is cumulative utility as a
    percentage of the 12-benchmark total. Four benchmarks reach 78.5\%.}
    \label{fig:rq3-selection}
\end{figure}

We start from $S = \emptyset$, where $R^2 = 0$ and the utility reduces to
discrimination alone, so the first pick is simply the benchmark that separates
models most sharply. We then repeatedly add $\arg\max_b U(b \mid S)$ and
recompute the utility of every remaining candidate against the enlarged set. We
run the path through all 12 benchmarks rather than stopping at a preset size,
so that the point at which the suite stops gaining is something we read off the
curve instead of fixing in advance (Figure~\ref{fig:rq3-selection}). At each
step we record the cumulative utility of the selected set and report it as a
percentage of the all-12 total, so that each benchmark's contribution can be
read relative to the whole suite rather than in absolute units of $U$.

Selection opens on \textsc{RefSpatial-Bench}, the suite's most discriminating
benchmark, then takes \textsc{MindCube}, \textsc{VSI-Bench}, and \textsc{BLINK}.
Those four reach 78.5\% of the utility of all 12. \textsc{Where2Place} carries
the second-highest Gini coefficient in Table~\ref{tab:benchmarks} yet the procedure passes
over it three times: with \textsc{RefSpatial-Bench} already selected, most of
its variance is reproducible, and the utility discounts it accordingly. This is
the substitute relationship of Section~\ref{sec:redundancy} acting exactly as
the utility intends, and it is the clearest illustration of why marginal
information cannot be judged benchmark by benchmark in isolation.
\textsc{Where2Place} enters fifth, at 85.0\% cumulative. The remaining seven
benchmarks share the last 15.0\%, and the final four add 4.4\% between them. We
therefore take the first four as the minimal suite.
Appendix~\ref{app:selection} gives further details on the forward selection and
tests the stability of the core against the opening pick.

The four benchmarks we obtain also cover complementary skills, which we did not
impose and which the utility has no way to encode: precise localization
(\textsc{RefSpatial-Bench}), consistency of a spatial model across limited views
(\textsc{MindCube}), spatial reasoning over video (\textsc{VSI-Bench}), and
multi-image perceptual primitives (\textsc{BLINK}). The core also balances the
two kinds of value the leave-one-out analysis distinguished. \textsc{MindCube}
is the suite's hub, the most frequent top predictor of the other benchmarks, so
its score carries information about the columns the core drops.
\textsc{BLINK} is the most isolated benchmark in the suite, and contributes
evidence no other benchmark supplies.

\subsection{Ranking Models}
\label{sec:rq3_ranking}

We rank models on the four selected benchmarks with a Bradley--Terry model fit in the
style of a preference arena \citep{chiang2024chatbot}, where each benchmark plays the
role of a judge. For every pair of models and every benchmark both have been scored on,
we record one binary observation: the benchmark votes for whichever model scored higher,
regardless of the size of the gap. Each vote carries equal weight, giving 4{,}231
observations over the 51 models. We then fit $P(i \succ j) = \sigma(r_i - r_j)$ by
maximum likelihood with an $L_2$ penalty of $10^{-3}$, center the strengths, and map them
to the familiar scale as $\text{Elo}_i = 1500 + 400\,r_i / \ln 10$
(Table~\ref{tab:elo}).

\begin{table}[t]
\centering
\caption{The compact leaderboard: top 10 of 51 models under a benchmark-as-judge
Bradley--Terry fit on the four selected benchmarks. \emph{Core} is how many of the four
the model has a score on.}
\label{tab:elo}
\vspace{0.5em}
\small
\begin{tabular}{clrc}
\toprule
\textbf{Rank} & \textbf{Model} & \textbf{Elo} & \textbf{Core} \\
\midrule
1  & \textsc{HY-Embodied-0.5 MoE-407B-A32B}     & 2251 & 3 \\
2  & \textsc{Qwen3.5-397B-A17B}                 & 2032 & 3 \\
3  & \textsc{Qwen3-VL-235B-A22B-Instruct}  & 1828 & 3 \\
4  & \textsc{Seed 2.0}                     & 1790 & 3 \\
5  & \textsc{Kimi K2.5}                    & 1750 & 4 \\
6  & \textsc{HY-Embodied-0.5 MoT-4B-A2B}       & 1744 & 4 \\
7  & \textsc{Gemini 3.0 Pro}               & 1716 & 4 \\
8  & \textsc{Qwen3-VL-32B-Instruct}        & 1662 & 3 \\
9  & \textsc{Gemini-2.5-Pro-preview-05-06} & 1643 & 3 \\
10 & \textsc{RoboBrain-32B-2.0}            & 1630 & 4 \\
\bottomrule
\end{tabular}
\end{table}

The leaderboard shows that top 10 models include both open weight and closed API systems. Three out of ten models in the top 10 (2 sizes of \textsc{HY-Embodied-0.5} and \textsc{RoboBrain-32B-2.0}) are models specifically post-trained for embodied or spatial tasks. If physical AI scores were driven purely
by general capability, the ranking would reproduce a general-purpose
leaderboard and domain post-training would provide no additional gain.

Interestingly, the leaderboard shows that the scale improves physical
ability within a training recipe but not between recipes. While
within the \textsc{Qwen-VL} family the ordering is monotone in size, leaderboard places \textsc{HY-Embodied-0.5 MoT-4B-A2B} sixth and above models significantly larger in terms of parameter count.

\section{Limitations}
\label{sec:limitations}

The insights gained from this research are subject to a few limitations that simultaneously point toward compelling directions for future study.

\paragraph{Benchmark-level analysis.} We operate on aggregate scores because that
is what the field publishes. Item-level audits can localize redundancy to specific items
and estimate measurement error directly \citep{metabench_2024}.

\paragraph{Matrix score verification and possible heterogeneity and noise.} Cells come from model cards,
benchmark papers, and our own runs. We aligned to official evaluation code and prompts
where available, but we have no systematic reproduction study and do not claim one. The usable
evidence on comparability is published-versus-published already revealed some inconsistencies. Moreover, a benchmark can be
unpredictable from its peers because it measures something distinct or because it is
noisy. Distinguishing the two requires repeated evaluation under resampled prompts,
decoding seeds, and parsing rules --- data current reporting does not provide.

\paragraph{Sample size.} We acknowledge that fifty-one models over 12 benchmarks matrix may be thin for some statistical analysis but try to rely on the findings supported by statistical significance.

\paragraph{The compact suite is one defensible choice, not the optimum.} Forward selection
is greedy and carries no optimality guarantee, and the recovered core depends on the
utility. All four slots survive when any core member is forced into the opening slot;
forcing a benchmark the unconstrained path rejects displaces one member to fifth and
lowers the utility captured at four (Appendix~\ref{app:selection}). We have not varied
the discrimination measure itself, so the core is robust to where selection starts but
untested against a different definition of $g$.

\paragraph{Observational evidence only.} Every result here is observational. We do not
intervene on training data or objectives, so we cannot claim the dominant axis
\emph{causes} performance on any benchmark, only that the two covary tightly across the
models that exist today. Finally, none of the 12 benchmarks measures downstream task
success on a physical system, so we cannot say whether the residual physical signal we
isolate is the part that transfers to manipulation or navigation.

\section{Conclusion}
\label{sec:conclusion}

We audited a 12-benchmark physical AI suite as a measurement instrument rather than a
scoreboard, using a matrix of 51 models assembled from published reports and our own
evaluation runs. The suite is substantially redundant: correlations are uniformly
positive, averaging 0.487, and roughly half of what the benchmarks share is general
vision--language capability (shared with general benchmark) rather than anything specific to physical AI
(Appendix~\ref{app:structure}). The redundancy is compressible: four benchmarks selected
for discrimination and marginal uniqueness retain 78.5\% of the suite's discriminating
power. The audit itself uses no property specific to physical AI: it needs only a set of benchmarks and
a set of models scored on enough of them to overlap and can be applied to any field. 

\bibliography{references}
\bibliographystyle{plainnat}

\appendix

\section{Complete Model List and Score Sources}
\label{app:models}

Table~\ref{tab:app-models} lists every model in the matrix with its provider, parameter
count, release date and base checkpoint, together with how its twelve scores were
obtained. 
Of the 564 filled cells, 405 are grouped as published: 379 are direct
transcriptions from a model card or paper, 25 are medians of conflicting
published values, and one is borrowed from a twin model. The remaining 159
cells are our own runs.

\footnotesize
\begin{longtable}{llrllrr}
\caption{The 51 models in the matrix: provider, parameter count (open-weight only), release date, base model where the checkpoint is a post-train of a named model, and how many of its 12 benchmark scores are grouped as published versus produced by our own runs.}\label{tab:app-models}\\
\toprule
Model & Provider & Size & Released & Base model & Pub. & Own \\
\midrule
\endfirsthead
\multicolumn{7}{l}{\footnotesize\itshape Table \thetable\ continued from the previous page.}\\
\toprule
Model & Provider & Size & Released & Base model & Pub. & Own \\
\midrule
\endhead
\midrule
\multicolumn{7}{r}{\footnotesize\itshape continued on the next page}\\
\endfoot
\bottomrule
\endlastfoot
Claude-Sonnet-4-2025-05-14 & Anthropic & --- & 2025-05 & --- & 10 & 0 \\
RoboBrain-32B-2.0 & BAAI & 32B & 2025-07 & Qwen2.5-VL-32B-Instruct & 8 & 3 \\
RoboBrain-7B-2.0 & BAAI & 7B & 2025-07 & Qwen2.5-VL-7B-Instruct & 9 & 3 \\
RoboBrain-7B-1.0 & BAAI & 7B & 2025-02 & LLaVA-OneVision-7B & 10 & 2 \\
RoboBrain-2.5-4B & BAAI & 4B & 2026-01 & Qwen3-VL-4B-Instruct & 8 & 3 \\
Seed 2.0 (ByteDance) & ByteDance & --- & 2026-01 & --- & 8 & 0 \\
Gemini 2.5 Pro & Google & --- & 2025-03 & --- & 11 & 1 \\
Gemini 2.5 Flash & Google & --- & 2025-05 & --- & 11 & 1 \\
Gemini Robotics-ER 1.5 & Google & --- & 2025-09 & gemini\_2\_5 & 10 & 0 \\
Gemini 3.0 Pro & Google & --- & 2025-11 & --- & 9 & 0 \\
Gemini-2.5-Pro-preview-05-06 & Google & --- & 2025-05 & --- & 9 & 0 \\
Gemini Robotics-ER (original) & Google & --- & 2025-03 & gemini\_2\_0 & 9 & 0 \\
HY-Embodied-0.5 MoT-4B-A2B & Tencent & 2B & 2026-04 & --- & 10 & 2 \\
HY-Embodied-0.5 MoE-407B-A32B & Tencent & --- & 2026-04 & --- & 8 & 0 \\
InternVL3.5-241B-A28B & Shanghai AI Lab & 241B & 2025-08 & --- & 9 & 0 \\
InternVL3.5-38B & Shanghai AI Lab & 38B & 2025-08 & --- & 9 & 0 \\
InternVL3.5-30B-A3B & Shanghai AI Lab & 30B & 2025-08 & --- & 5 & 7 \\
InternVL3.5-20B-A4B & Shanghai AI Lab & 20B & 2025-08 & --- & 5 & 7 \\
InternVL3.5-14B & Shanghai AI Lab & 14B & 2025-08 & --- & 5 & 7 \\
InternVL3.5-8B & Shanghai AI Lab & 8B & 2025-08 & --- & 5 & 7 \\
InternVL2-8B & Shanghai AI Lab & 8B & 2024-07 & --- & 4 & 7 \\
InternVL3.5-4B & Shanghai AI Lab & 4B & 2025-08 & --- & 5 & 7 \\
InternVL2-2B & Shanghai AI Lab & 2B & 2024-07 & --- & 3 & 9 \\
InternVL3.5-2B & Shanghai AI Lab & 2B & 2025-08 & --- & 5 & 6 \\
InternVL3.5-1B & Shanghai AI Lab & 1B & 2025-08 & --- & 5 & 7 \\
Kimi K2.5 & Moonshot & --- & 2026-02 & --- & 9 & 3 \\
LLaVA-OneVision-7B & LLaVA & 7B & 2024-08 & --- & 3 & 9 \\
VeBrain-8B & Meta & 8B & 2025-05 & Qwen2.5-VL-7B-Instruct & 9 & 3 \\
Cosmos-Reason2-8B & NVIDIA & 8B & 2026-04 & Qwen3-VL-8B-Instruct & 4 & 8 \\
Magma-8B & NVIDIA & 8B & 2025-02 & --- & 9 & 3 \\
Cosmos-Reason1-7B & NVIDIA & 7B & 2025-03 & Qwen2.5-VL-7B-Instruct & 9 & 3 \\
Cosmos-Reason2-2B & NVIDIA & 2B & 2026-04 & Qwen3-VL-2B-Instruct & 4 & 8 \\
GPT-4o-2024-11-20 & OpenAI & --- & 2024-11 & --- & 11 & 1 \\
GPT-5-mini & OpenAI & --- & 2025-08 & GPT-5 & 11 & 1 \\
GPT-5 & OpenAI & --- & 2025-08 & --- & 11 & 1 \\
GPT-5.4 & OpenAI & --- & --- & --- & 8 & 1 \\
GPT-o4-mini-2025-05-16 & OpenAI & --- & 2025-05 & --- & 8 & 0 \\
Qwen3-VL-235B-A22B-Instruct & Alibaba & 235B & 2025-09 & --- & 7 & 4 \\
Qwen2.5-VL-72B-Instruct & Alibaba & 72B & 2025-01 & --- & 11 & 1 \\
Qwen3-VL-32B-Instruct & Alibaba & 32B & 2025-09 & --- & 4 & 7 \\
Qwen2.5-VL-32B-Instruct & Alibaba & 32B & 2025-01 & --- & 11 & 0 \\
Qwen3-VL-8B-Instruct & Alibaba & 8B & 2025-09 & --- & 7 & 5 \\
Qwen2.5-VL-7B-Instruct & Alibaba & 7B & 2025-01 & --- & 12 & 0 \\
Qwen3-VL-4B-Instruct & Alibaba & 4B & 2025-09 & --- & 8 & 3 \\
Qwen2.5-VL-3B & Alibaba & 3B & 2025-01 & --- & 5 & 7 \\
Qwen3-VL-2B-Instruct & Alibaba & 2B & 2025-09 & --- & 10 & 2 \\
Qwen3.5-397B-A17B & Alibaba & --- & 2026-02 & --- & 8 & 3 \\
Embodied-R1.5 & Tianjin Univ. & 8B & 2026 & Qwen3-VL-8B-Instruct & 9 & 2 \\
Embodied-R1 & Tianjin Univ. & --- & 2025 & Qwen2.5-VL-3B & 8 & 3 \\
Pelican-VL & X Humanoid & --- & --- & --- & 10 & 0 \\
MiMo-Embodied-7B & Xiaomi & 7B & 2025-12 & --- & 9 & 2 \\
\end{longtable}

\section{Structure: How the Physical Suite Relates to General Capability}
\label{app:structure}

Section~\ref{sec:redundancy} establishes that the 12 benchmarks share a great deal of
variance but does not qualify what that shared variance is. The
benchmarks may share a physical competence or they may share general model capability, in which case the suite
is a vision--language leaderboard wearing a physical label. The two are not
distinguishable from inside the suite, because any axis estimated from the 12 benchmarks
inherits whatever they have in common. This appendix separates them using evidence the
suite does not contain.

A set of 9 general language and vision benchmarks not
designed to measure spatial or 3D reasoning are added to the main matrix for this analysis: \textsc{MMMU}~\citep{yue2023mmmu}
and \textsc{MMStar}~\citep{chen2024mmstar} (multi-discipline multimodal reasoning),
\textsc{MMLU-Pro}~\citep{wang2024mmlupro} and
\textsc{GPQA-Diamond}~\citep{rein2023gpqa} (text-only knowledge and graduate-level
reasoning), \textsc{DocVQA}~\citep{mathew2021docvqa} and
\textsc{OCRBench}~\citep{liu2024ocrbench} (document and text-in-image reading),
\textsc{Video-MME}~\citep{fu2024videomme} (general video understanding), and the
text~\citep{chiang2024chatbot} and vision~\citep{chou2024visionarena}
\textsc{LM-Arena} Elo ratings (human preference). They were
chosen on two criteria: they are not intended to test physical or spatial understanding,
and they are widely reported for the same models that populate the physical-AI matrix.

Before the general anchors enter the analysis, it is important to understand the structure of the physical benchmarks matrix. PCA on the matrix of the 12
z-scored benchmarks yields a first component explaining 55.2\% of total variance, and
every benchmark loads on it positively --- from $0.48$ (\textsc{RealWorldQA}) to $0.88$
(\textsc{CV-Bench}). 

The general anchors are held out of the PCA entirely: PC1 is estimated from the 12 physical benchmarks alone and
only then correlated against general performance, so no anchor can influence the
component it is being compared to. The alignment turns out to be near-complete
(Table~\ref{tab:app-anchors}). Physical PC1 tracks the
first principal component of the anchors at Spearman $\rho = 0.952$, and it reaches
$0.950$ against \textsc{MMStar} and $0.942$ against \textsc{Video-MME}. In other words, a model's
score on a document-reading benchmark predicts its position on the physical suite's
principal axis about as well as the physical benchmarks predict one another.

\begin{table}[h]
\centering
\caption{Spearman correlation between the physical suite's PC1 --- estimated from the 12
physical benchmarks alone --- and each general anchor, over the models reporting both.
$n$ = overlapping models; CI = bootstrap interval with the PCA refit inside each
resample. \emph{General PC1} is a
separate PC1-physical-vs-PC1-general comparison.}
\label{tab:app-anchors}
\vspace{0.5em}
\small
\begin{tabular}{lrrcc}
\toprule
\textbf{General anchor} & \textbf{$n$} & \textbf{$\rho$} & \textbf{95\% CI} \\
\midrule
\textsc{MMStar}            & 29 & 0.950 & $[0.85,\ 0.98]$ \\
\textsc{Video-MME}         & 26 & 0.942 & $[0.81,\ 0.99]$ \\
\textsc{MMLU-Pro}          & 16 & 0.871 & $[0.59,\ 0.98]$ \\
\textsc{MMMU}              & 37 & 0.820 & $[0.58,\ 0.95]$ \\
\textsc{OCRBench}          & 27 & 0.774 & $[0.46,\ 0.94]$ \\
\textsc{DocVQA}            & 22 & 0.754 & $[0.42,\ 0.93]$  \\
\textsc{LM-Arena (vision)}$^{\dagger}$ & 12 & 0.741 & $[0.34,\ 0.95]$ \\
\textsc{GPQA-Diamond}      & 23 & 0.733 & $[0.38,\ 0.92]$ \\
\textsc{LM-Arena (text)}$^{\dagger}$   & 10 & 0.624 & $[-0.14,\ 0.97]$ \\
\midrule
\textsc{General PC1}       & 32 & 0.952 & $[0.82,\ 0.97]$  \\
\bottomrule
\end{tabular}
\\[2pt]
{\footnotesize $^{\dagger}$ underpowered ($n < 15$).}
\end{table}

That result is about the suite as a whole. It does not say which individual pairs of
benchmarks agree because both track general capability, and which agree for some other
reason. To separate the two, we residualize each benchmark
on an externally estimated general axis --- the first PC of the 9 anchors --- and recompute the correlations among the 12
residual vectors. Removing that single external axis halves the suite's internal structure:
mean pairwise $|\rho|$ falls from $0.487$ to $0.250$, and the number of pairs correlating
above $0.5$ drops from 34 of 66 to 6. Roughly half of all benchmark-to-benchmark
agreement in physical AI is general capability. Three bonds
survive at nearly full strength and are therefore specific rather than general
(Table~\ref{tab:app-residual}): the pointing pair, the 2D-perception pair, and
\textsc{ERQA}\,$\leftrightarrow$\,\textsc{MindCube}. The first two are exactly the
substitute pairs flagged in Section~\ref{sec:redundancy}: they duplicate each other because they measure the same specific ability, not
because both are loaded on general capability. Pairs that looked substantial in RQ1 but
were mostly general capability collapse instead. Another findings is that \textsc{VSI-Bench} and \textsc{BLINK} are uncorrelated
in raw scores ($\rho = 0.002$) but correlate at $-0.363$ in the residuals, so among
models of equal general capability, strength in video-spatial reasoning trades off
against strength in multi-image perception. 

\begin{table}[h]
\centering
\caption{Benchmark-pair correlations before and after removing an externally estimated
general axis (the first PC of the 9 anchors). \emph{Residual} = correlation of the two
benchmarks' residuals, all 51 models; \emph{Dense block} = the same using only the 32
models with $\ge 4$ observed anchors and no imputation. Top: the three bonds that
survive. Middle: the \textsc{VSI}/\textsc{BLINK} suppression, which appears only after
conditioning. Bottom: two representative collapses.}
\label{tab:app-residual}
\vspace{0.5em}
\small
\begin{tabular}{llrrr}
\toprule
\textbf{Benchmark A} & \textbf{Benchmark B} & \textbf{Raw $\rho$} & \textbf{Residual $\rho$} & \textbf{Dense block} \\
\midrule
\textsc{EmbSpatial}  & \textsc{CV-Bench}         & 0.876 & \textbf{0.784} & 0.836 \\
\textsc{Where2Place} & \textsc{RefSpatial-Bench} & 0.860 & \textbf{0.801} & 0.773 \\
\textsc{ERQA}        & \textsc{MindCube}         & 0.656 & \textbf{0.613} & 0.658 \\
\midrule
\textsc{VSI-Bench}   & \textsc{BLINK}            & 0.002 & $-0.363$ & $-0.425$ \\
\midrule
\textsc{ERQA}        & \textsc{EmbSpatial}       & 0.729 & 0.197 & $-0.095$ \\
\textsc{ERQA}        & \textsc{CV-Bench}         & 0.725 & 0.243 & $-0.055$ \\
\bottomrule
\end{tabular}
\end{table}

\section{Forward Selection Details}
\label{app:selection}

This appendix supports Section~\ref{sec:rq3_suite}. It reports the full 12-step
selection path behind the compact suite, and then asks whether that suite is an artifact
of the one arbitrary choice greedy selection makes: its opening pick. 

Table~\ref{tab:app-path} gives all 12 steps of the greedy
selection under $U(b \mid S) = g(b)\,(1 - R^2(b \sim S))$. Marginal utility decays sharply --- the last four benchmarks contribute
4.4\% of the total between them. 

\begin{table}[h]
\centering
\caption{Full greedy selection path. $g$ is the Gini coefficient of the raw score spread,
$R^2$ is the fit of the candidate on the already-selected set, $U$ their product, and
\emph{Cum.} the running total as a percentage of the all-12 sum. $n$ is the number of
models on which $R^2$ is estimated.}
\label{tab:app-path}
\vspace{0.5em}
\small
\begin{tabular}{clrrrrr}
\toprule
\textbf{Step} & \textbf{Benchmark} & \textbf{$g$} & \textbf{$R^2$} & \textbf{$U$} & \textbf{Cum.\ \%} & \textbf{$n$} \\
\midrule
1  & \textsc{RefSpatial-Bench} & 0.343 & 0.000 & 0.3431 & 38.8  & 51 \\
2  & \textsc{MindCube}         & 0.153 & 0.087 & 0.1396 & 54.6  & 42 \\
3  & \textsc{VSI-Bench}        & 0.153 & 0.254 & 0.1144 & 67.5  & 42 \\
4  & \textsc{BLINK}            & 0.106 & 0.087 & 0.0972 & 78.5  & 33 \\
5  & \textsc{Where2Place}      & 0.265 & 0.783 & 0.0576 & 85.0  & 32 \\
6  & \textsc{RealWorldQA}      & 0.069 & 0.442 & 0.0383 & 89.4  & 28 \\
7  & \textsc{SAT}              & 0.097 & 0.668 & 0.0323 & 93.0  & 28 \\
8  & \textsc{OmniSpatial}      & 0.071 & 0.673 & 0.0232 & 95.6  & 27 \\
9  & \textsc{RoboSpatial}      & 0.103 & 0.816 & 0.0189 & 97.8  & 27 \\
10 & \textsc{EmbSpatial}       & 0.057 & 0.824 & 0.0100 & 98.9  & 27 \\
11 & \textsc{ERQA}             & 0.104 & 0.946 & 0.0056 & 99.5  & 27 \\
12 & \textsc{CV-Bench}         & 0.039 & 0.895 & 0.0041 & 100.0 & 27 \\
\bottomrule
\end{tabular}
\end{table}

Greedy selection commits to its first pick, so we force different benchmarks into the
opening slot and let the argmax proceed from there (Table~\ref{tab:app-seeds}). Forcing
a member of the core returns the identical set of four in a different order. Forcing
\textsc{Where2Place} --- the substitute of the natural opening pick --- does not keep
\textsc{RefSpatial-Bench} out: even with 75\% of its variance already reproducible from
the first three selections, its dispersion buys it back at step~4
($U = 0.343 \times 0.248 = 0.085$), so the path carries both pointing benchmarks and
\textsc{BLINK} is the member displaced to fifth.  Forcing a benchmark the unconstrained
path rejects (\textsc{RoboSpatial}, \textsc{SAT}) likewise displaces one core member to
fifth place but leaves the other three standing. Every forced start lowers the utility
captured at four benchmarks, and the ordering is informative: seeding with a core member
costs almost nothing (78.5 to 77.8--78.3), seeding with the substitute of a core member
costs about four points (74.5), and seeding with a rejected benchmark costs seven to
nine (69.2--71.6). \textsc{MindCube} and \textsc{VSI-Bench} enter within the first four
in every run, and \textsc{RefSpatial-Bench} is never kept out --- not even by its own
substitute. Physical specificity alone does not earn a place: \textsc{RoboSpatial}
carries the least general-capability signal in the suite (Appendix~\ref{app:structure}
finds it the benchmark least aligned with the suite's general-capability axis), yet its own dispersion
($g = 0.103$) is too low for it to anchor a ranking.

\begin{table}[h]
\centering
\caption{Seed robustness of the selection path. \emph{Kept} counts how many of the four
unseeded core benchmarks survive in the first four slots; \emph{Cum.\ \% @4} is the
utility captured at four benchmarks.}
\label{tab:app-seeds}
\vspace{0.5em}
\small
\begin{tabular}{llcr}
\toprule
\textbf{Forced first} & \textbf{First four selected} & \textbf{Kept} & \textbf{Cum.\ \% @4} \\
\midrule
\emph{(none)}        & RefSpatial-Bench, MindCube, VSI-Bench, BLINK & 4/4 & 78.5 \\
\textsc{VSI-Bench}   & VSI-Bench, RefSpatial-Bench, MindCube, BLINK & 4/4 & 77.8 \\
\textsc{BLINK}       & BLINK, RefSpatial-Bench, MindCube, VSI-Bench & 4/4 & 78.3 \\
\textsc{Where2Place} & Where2Place, MindCube, VSI-Bench, RefSpatial-Bench & 3/4 & 74.5 \\
\textsc{RoboSpatial} & RoboSpatial, RefSpatial-Bench, MindCube, VSI-Bench & 3/4 & 69.2 \\
\textsc{SAT}         & SAT, RefSpatial-Bench, VSI-Bench, MindCube & 3/4 & 71.6 \\
\bottomrule
\end{tabular}
\end{table}

Forward selection is greedy and carries no optimality guarantee, and the ablation above
tests only one of its degrees of freedom. The core should therefore be considered as a compact suite under this utility, not as the unique optimum.
\end{document}